# Deep Learning in Bioinformatics

Seonwoo Min[1], Byunghan Lee[1], and Sungroh Yoon[1,2*]

[1]Department of Electrical and Computer Engineering, Seoul National University, Seoul 08826, Korea

[2]Interdisciplinary Program in Bioinformatics, Seoul National University, Seoul 08826, Korea

**Abstract**

In the era of big data, transformation of biomedical big data into valuable knowledge has been one of the most important challenges in bioinformatics. Deep learning has advanced rapidly since the early 2000s and now demonstrates state-of-the-art performance in various fields. Accordingly, application of deep learning in bioinformatics to gain insight from data has been emphasized in both academia and industry. Here, we review deep learning in bioinformatics, presenting examples of current research. To provide a useful and comprehensive perspective, we categorize research both by the bioinformatics domain (*i.e.*, omics, biomedical imaging, biomedical signal processing) and deep learning architecture (*i.e.*, deep neural networks, convolutional neural networks, recurrent neural networks, emergent architectures) and present brief descriptions of each study. Additionally, we discuss theoretical and practical issues of deep learning in bioinformatics and suggest future research directions. We believe that this review will provide valuable insights and serve as a starting point for researchers to apply deep learning approaches in their bioinformatics studies.

[*]Corresponding author. Mailing address: 301-908, Department of Electrical and Computer Engineering, Seoul National University, Seoul 08826, Korea. E-mail: sryoon@snu.ac.kr. Phone: +82-2-880-1401.

## Key Points

- As a great deal of biomedical data have been accumulated, various machine algorithms are now being widely applied in bioinformatics to extract knowledge from big data.

- Deep learning, which has evolved from the acquisition of big data, the power of parallel and distributed computing, and sophisticated training algorithms, has facilitated major advances in numerous domains such as image recognition, speech recognition, and natural language processing.

- We review deep learning for bioinformatics and present research categorized by bioinformatics domain (*i.e.*, omics, biomedical imaging, biomedical signal processing) and deep learning architecture (*i.e.*, deep neural networks, convolutional neural networks, recurrent neural networks, emergent architectures).

- Furthermore, we discuss the theoretical and practical issues plaguing the applications of deep learning in bioinformatics, including imbalanced data, interpretation, hyperparameter optimization, multimodal deep learning, and training acceleration.

- As a comprehensive review of existing works, we believe that this paper will provide valuable insight and serve as a launching point for researchers to apply deep learning approaches in their bioinformatics studies.

# Author Description

**Seonwoo Min** is a M.S./Ph.D. candidate at Department of Electrical and Computer Engineering, Seoul National University, Korea. His research areas include high-performance bioinformatics, machine learning for biomedical big data, and deep learning.

**Byunghan Lee** is a Ph.D. candidate at Department of Electrical and Computer Engineering, Seoul National University, Korea. His research areas include high-performance bioinformatics, machine learning for biomedical big data, and data mining.

**Sungroh Yoon** is an associate professor at Department of Electrical and Computer Engineering, Seoul National University, Seoul, Korea. He received his Ph.D. and postdoctoral training from Stanford University, Stanford, USA. His research interests include machine learning and deep learning for bioinformatics, and high-performance bioinformatics.

# Introduction

**TABLE 1**: Abbreviations in alphabetical order

| Abbreviation | Full word |
|---|---|
| AE | Auto-Encoder |
| AI | Artificial Intelligence |
| AUC | Area Under the Receiver Operation Characteristics Curve |
| AUC-PR | Area Under the Precision-Recall Curve |
| BRNN | Bidirectional Recurrent Neural Network |
| CAE | Convolutional Auto-Encoder |
| CNN | Convolutional Neural Network |
| DBN | Deep Belief Network |
| DNN | Deep Neural Network |
| DST-NN | Deep Spatio-Temporal Neural Network |
| ECG | Electrocardiography |
| ECoG | Electrocorticography |
| EEG | Electroencephalography |
| EMG | Electromyography |
| EOG | Electrooculography |
| GRU | Gated Recurrent Unit |
| LSTM | Long Short-Term Memory |
| MD-RNN | Multi-dimensional Recurrent Neural Network |
| MLP | Multilayer Perceptron |
| MRI | Magnetic Resonance Image |
| PCA | Principal Component Analysis |
| PET | Positron Emission Tomography |
| PSSM | Position Specific Scoring Matrix |
| RBM | Restricted Boltzmann Machine |
| ReLU | Rectified Linear Unit |
| RNN | Recurrent Neural Network |
| SAE | Stacked Auto-Encoder |
| SGD | Stochastic Gradient Descent |

In the era of "big data," transformation of large quantities of data into valuable knowledge has become increasingly important in various domains [1], and bioinformatics is no exception. Significant amounts of biomedical data, including omics, image, and signal data, have been accumulated, and the resulting potential for applications in biological and healthcare research has caught the attention of both industry and academia. For instance, IBM developed Watson for Oncology, a platform analyzing patients' medical information and assisting clinicians with treatment options [2, 3]. In addition, Google DeepMind, having achieved great success with AlphaGo in the game of Go, recently launched DeepMind Health to develop effective healthcare technologies [4, 5].

To extract knowledge from big data in bioinformatics, machine learning has been a widely used and successful methodology. Machine learning algorithms use training data to uncover underlying patterns, build models, and make predictions based on the best fit model. Indeed, some well-known algorithms (*i.e.,* support vector machines, random forests, hidden Markov models, Bayesian networks, Gaussian networks) have been applied in genomics, proteomics, systems biology, and numerous other domains [6].

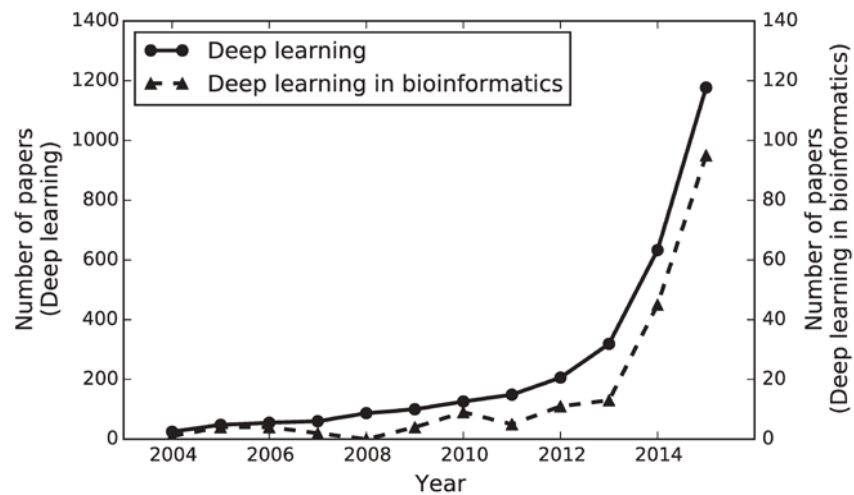

[FIGURE 1]

The proper performance of conventional machine learning algorithms relies heavily on data representations called features [7]. However, features are typically designed by human engineers with extensive domain expertise, and identifying which features are more appropriate for the given task remains difficult. Deep learning, a branch of machine learning, has recently emerged based on big data, the power of parallel and distributed computing, and sophisticated algorithms. Deep learning has overcome previous limitations, and academic interest has increased rapidly since the early 2000s (Figure 1). Furthermore deep learning is responsible for major advances in diverse fields where the artificial intelligence (AI) community has struggled for many years [8]. One of the most important advancements thus far has been in image and speech recognition [9-15], although promising results have been disseminated in natural language processing [16, 17] and language translation [18, 19]. Certainly, bioinformatics can also benefit from deep learning (Figure 2): splice junctions can be discovered from DNA sequences, finger joints can be recognized from X-ray images, lapses can be detected from electroencephalography (EEG) signals, and so on.

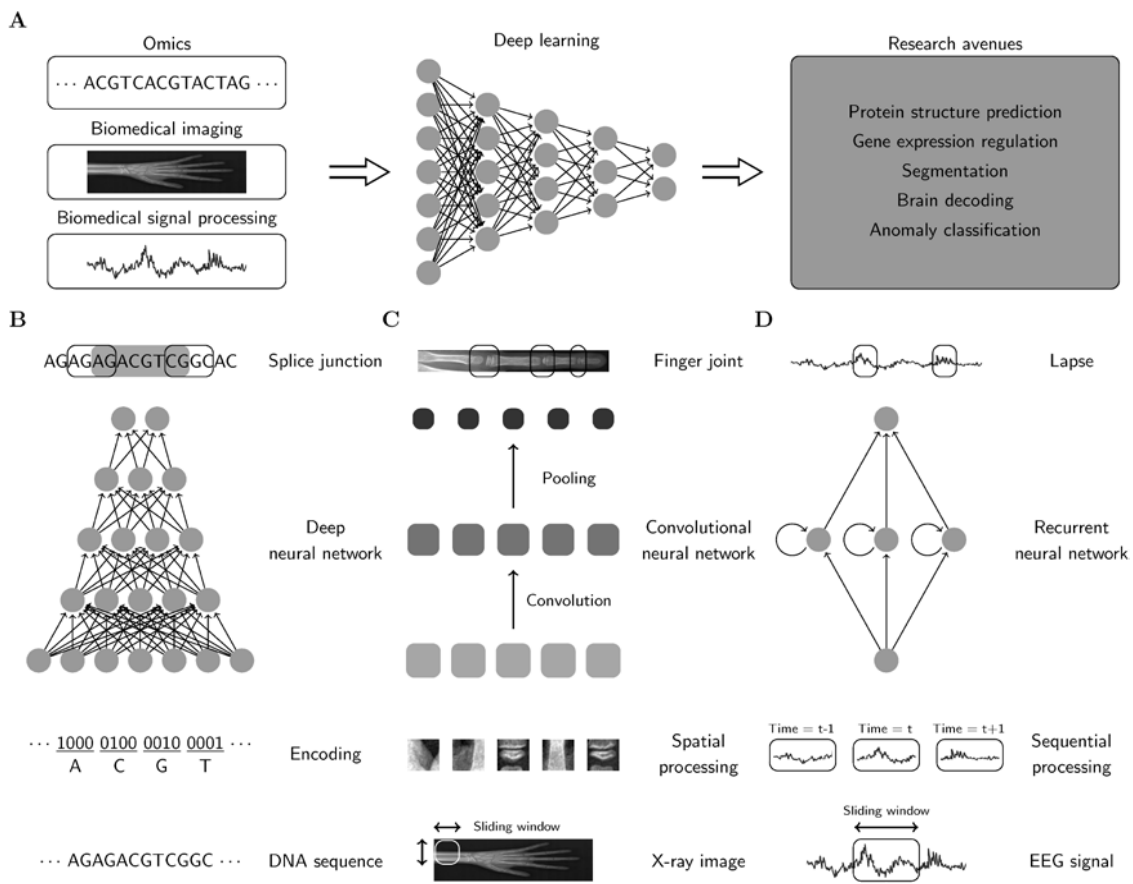

[FIGURE 2]

Previous reviews have addressed machine learning in bioinformatics [6, 20] and the fundamentals of deep learning [7, 8, 21]. In addition, although recently published reviews by Leung et al. [22], Mamoshina et al. [23], and Greenspan et al. [24] discussed deep learning applications in bioinformatics research, the former two are limited to applications in genomic medicine, and the latter to medical imaging. In this article, we provide a more comprehensive review of deep learning for bioinformatics and research examples categorized by bioinformatics domain (*i.e.*, omics, biomedical imaging, biomedical signal processing) and deep learning architecture (*i.e.*, deep neural networks, convolutional neural networks, recurrent neural networks, emergent architectures). The goal of this article is to provide valuable insight and to serve as a starting point to facilitate the application of deep learning in bioinformatics studies. To the best of our knowledge, we are one of the first groups to review deep learning applications in bioinformatics.

# Deep learning: a brief overview

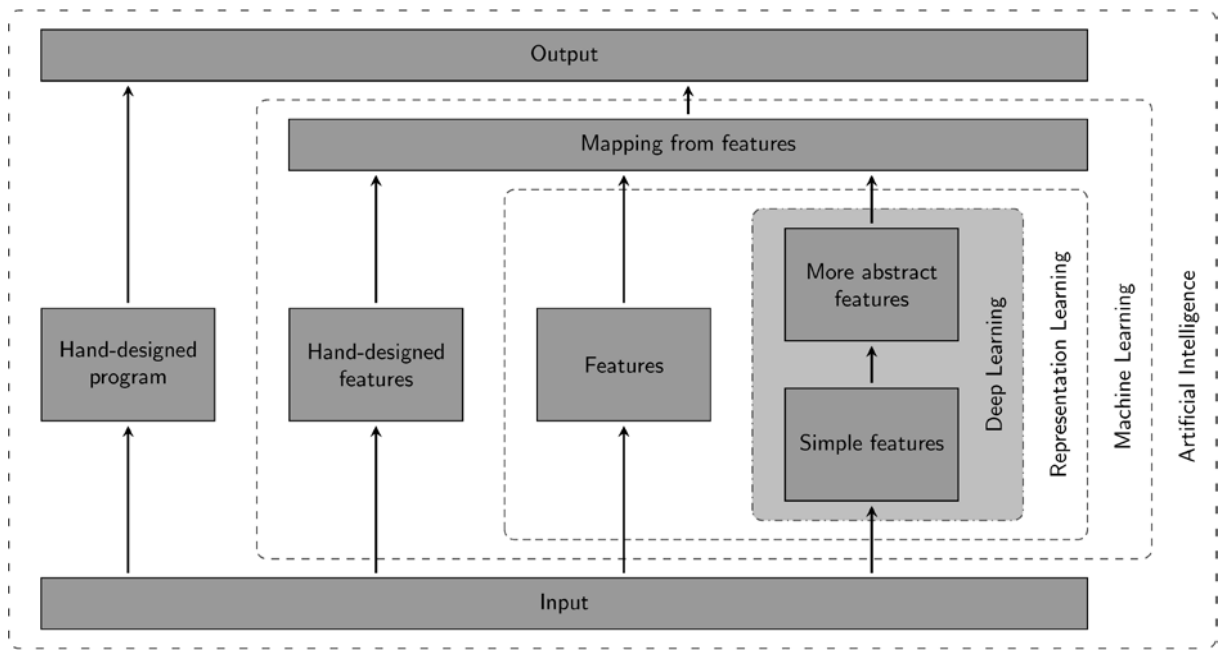

[FIGURE 3]

Efforts to create AI systems have a long history. Figure 3 illustrates the relationships and high-level schematics of different disciplines. Early approaches attempted to explicitly program the required knowledge for given tasks; however, these faced difficulties in dealing with complex real-world problems because designing all the detail required for an AI system to accomplish satisfactory results by hand is such a demanding job [7]. Machine learning provided more viable solutions with the capability to improve through experience and data. Although machine learning can extract patterns from data, there are limitations in raw data processing, which is highly dependent on hand-designed features. To advance from hand-designed to data-driven features, representation learning, particularly deep learning has shown great promise. Representation learning can discover effective features as well as their mappings from data for given tasks. Furthermore, deep learning can learn complex features by combining simpler features learned from data. In other words, with artificial neural networks of multiple nonlinear layers, referred to as deep learning architectures, hierarchical representations of data can be discovered with increasing levels of abstraction [25].

## Key elements of deep learning

The successes of deep learning are built on a foundation of significant algorithmic details and generally can be understood in two parts: construction and training of deep learning architectures. Deep learning architectures are basically artificial neural networks of multiple nonlinear layers and several types have been proposed according to input data characteristics and research objectives. Here, we categorized deep learning architectures into four groups (*i.e.,* deep neural networks (DNNs) [26-30], convolutional neural networks (CNNs) [31-33], recurrent neural networks (RNNs) [34-37], emergent architectures [38-41]) and explained each group in detail (Table 2). Some papers have used "DNNs" to encompass all deep learning architectures [7, 8]; however, in this review, we use "DNNs" to refer specifically to multilayer perceptron (MLP) [26], stacked auto-encoder (SAE) [27, 28], and deep belief networks (DBNs) [29, 30], which use perceptrons [42], auto-encoders (AEs) [43], and restricted Boltzmann machines (RBMs) [44, 45] as the building blocks of neural networks, respectively. CNNs are architectures that have succeeded particularly in image recognition and consist of convolution layers, nonlinear layers, and pooling layers. RNNs are designed to utilize sequential information of input data with cyclic connections among building blocks like perceptrons, long short-term memory units (LSTMs) [36, 37], or gated recurrent units (GRUs) [19]. In addition, many other emergent deep learning architectures have been suggested, such as deep spatio-temporal neural networks (DST-NNs) [38], multi-dimensional recurrent neural networks (MD-RNNs) [39], and convolutional auto-encoders (CAEs) [40, 41].

**TABLE 2**: Categorization of deep learning applied research in bioinformatics

|  | Omics | | Biomedical imaging | | Biomedical signal processing | |
| --- | --- | --- | --- | --- | --- | --- |
|  | Research topics | Reference | Research topics | Reference | Research topics | Reference |
| **Deep neural networks** | Protein structure | [84-87] | Anomaly classification | [122-124] | Brain decoding | [158-163] |
|  | Gene expression regulation | [93-98] | Segmentation | [133] | Anomaly classification | [171-175] |
|  | Protein classification | [108] | Recognition | [142, 143] | | |
|  | Anomaly classification | [111] | Brain decoding | [149, 150] | | |
| **Convolutional neural networks** | Gene expression regulation | [99-104] | Anomaly classification | [125-132] | Brain decoding | [164-167] |
|  | | | Segmentation | [134-140] | Anomaly classification | [176] |
|  | | | Recognition | [144-147] | | |
| **Recurrent neural networks** | Protein structure | [88-90] | | | Brain decoding | [168] |
|  | Gene expression regulation | [105-107] | | | Anomaly classification | [177, 178] |
|  | Protein classification | [109, 110] | | | | |
| **Emergent architectures** | Protein structure | [91, 92] | Segmentation | [141] | Brain decoding | [169, 170] |

The goal of training deep learning architectures is optimization of the weight parameters in each layer, which gradually combines simpler features into complex features so that the most suitable hierarchical representations can be learned from data. A single cycle of the optimization process is organized as follows [8]. First, given a training dataset, the forward pass sequentially computes the output in each layer and propagates the function signals forward through the network. In the final output layer, an objective loss function measures error between the inferenced outputs and the given labels. To minimize the training error, the backward pass uses the chain rule to backpropagate error signals and compute gradients with respect to all weights throughout the neural network [46]. Finally, the weight parameters are updated using optimization algorithms based on stochastic gradient descent (SGD) [47]. Whereas batch gradient descent performs parameter updates for each complete dataset, SGD provides stochastic approximations by performing the updates for each small set of data examples. Several optimization algorithms stem from SGD. For example, Adagrad [48] and Adam [49] perform SGD while adaptively modifying learning rates based on update frequency and moments of the gradients for each parameter, respectively.

Another core element in the training of deep learning architectures is regularization, which refers to strategies intended to avoid overfitting and thus achieve good generalization performance. For example, weight decay [50], a well-known conventional approach, adds a penalty term to the objective loss function so that weight parameters converge to smaller absolute values. Currently, the most widely used regularization approach is dropout [51]. Dropout randomly removes hidden units from neural networks during training and can be considered an ensemble of possible subnetworks [52]. To enhance the capabilities of dropout, a new activation function, maxout [53], and a variant of dropout for RNNs called rnnDrop [54], have been proposed. Furthermore, recently proposed batch normalization [55] provides a new regularization method through normalization of scalar features for each activation within a mini-batch and learning each mean and variance as parameters.

**Deep learning libraries**

To actually implement deep learning algorithms, a great deal of attention to algorithmic details is required. Fortunately, many open source deep learning libraries are available online (Table 3). There are still no clear front-runners, and each library has its own strengths [56]. According to benchmark test results of CNNs, specifically AlexNet [33] implementation in Baharampour et al. [57], Python-based Neon [58] shows a great advantage in the processing speed. C++ based Caffe [59] and Lua-based Torch [60] offer great advantages in terms of pre-trained models and functional extensionality, respectively. Python-based Theano [61, 62] provides a low-level library to define and optimize mathematical expressions; moreover, numerous higher-level wrappers such as Keras [63], Lasagne [64], and Blocks [65] have been developed on top of Theano to provide more intuitive interfaces. Google recently released the C++-based TensorFlow [66] with a Python interface. This library currently shows limited performance but is undergoing continuous improvement, as heterogeneous distributed computing is now supported. In addition, TensorFlow can also take advantage of Keras, which provides an additional model-level interface.

**TABLE 3**: Comparison of deep learning libraries

|  | Core | Speed for batch* (ms) | Multi-GPU | Distributed | Strengths [56, 57] |
|---|---|---|---|---|---|
| **Caffe** | C++ | 651.6 | O | X | Pre-trained models supported |
| **Neon** | Python | 386.8 | O | X | Speed |
| **TensorFlow** | C++ | 962.0 | O | O | Heterogeneous distributed computing |
| **Theano** | Python | 733.5 | X | X | Ease of use with higher-level wrappers |
| **Torch** | Lua | 506.6 | O | X | Functional extensionality |

*Notes.* Speed for batch* is based on the averaged processing times for AlexNet [33] with batch size of 256 on a single GPU [57];
Caffe, Neon, Theano, Torch was utilized with cuDNN v.3 while TensorFlow was utilized with cuDNN v.2

# Deep neural networks

The basic structure of DNNs consists of an input layer, multiple hidden layers, and an output layer (Figure 4). Once input data are given to the DNNs, output values are computed sequentially along the layers of the network. At each layer, the input vector comprising the output values of each unit in the layer below is multiplied by the weight vector for each unit in the current layer to produce the weighted sum. Then, a nonlinear function, such as a sigmoid, hyperbolic tangent, or rectified linear unit (ReLU) [67], is applied to the weighted sum to compute the output values of the layer. The computation in each layer transforms the representations in the layer below into slightly more abstract representations [8]. Based on the types of layers used in DNNs and the corresponding learning method, DNNs can be classified as MLP, SAE, or DBN.

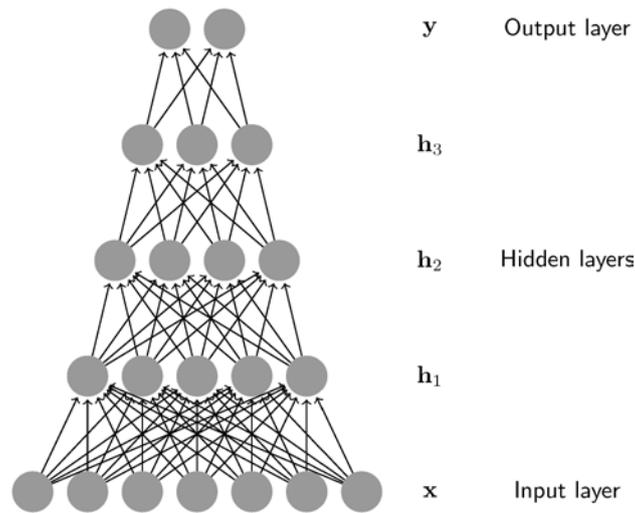

[FIGURE 4]

MLP has a similar structure to the usual neural networks but includes more stacked layers. It is trained in a purely supervised manner that uses only labeled data. Since the training method is a process of optimization in high-dimensional parameter space, MLP is typically used when a large number of labeled data are available [25].

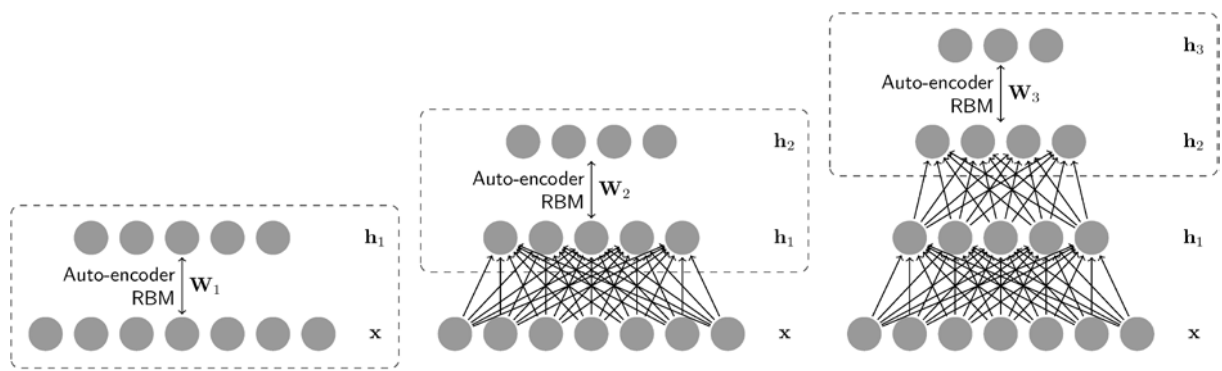

[FIGURE 5]

SAE and DBN use AEs and RBMs as building blocks of the architectures, respectively. The main difference between these and MLP is that training is executed in two phases: unsupervised pre-training and supervised fine-tuning. First, in unsupervised pre-training (Figure 5), the layers are stacked sequentially and trained in a layer-wise manner as an AE or RBM using unlabeled data. Afterwards, in supervised fine-tuning, an output classifier layer is stacked, and the whole neural network is optimized by retraining with labeled data. Since both SAE and DBN exploit unlabeled data and can help avoid overfitting, researchers are able to obtain fairly regularized results, even when labeled data are insufficient as is common in the real world [68].

DNNs are renowned for their suitability in analyzing high-dimensional data. Given that bioinformatics data are typically complex and high-dimensional, DNNs have great promise for bioinformatics research. We believe DNNs, as hierarchical representation learning methods, can discover previously unknown highly abstract patterns and correlations to provide insight to better understand the nature of the data. However, it has occurred to us that the capabilities of DNNs have not yet fully been exploited. Although the key characteristic of DNNs is that hierarchical features are learned solely from data, human designed features have often been given as inputs instead of raw data forms. We expect that the future progress of DNNs in bioinformatics will come from investigations into proper ways to encode raw data and learn suitable features from them.

## Convolutional neural networks

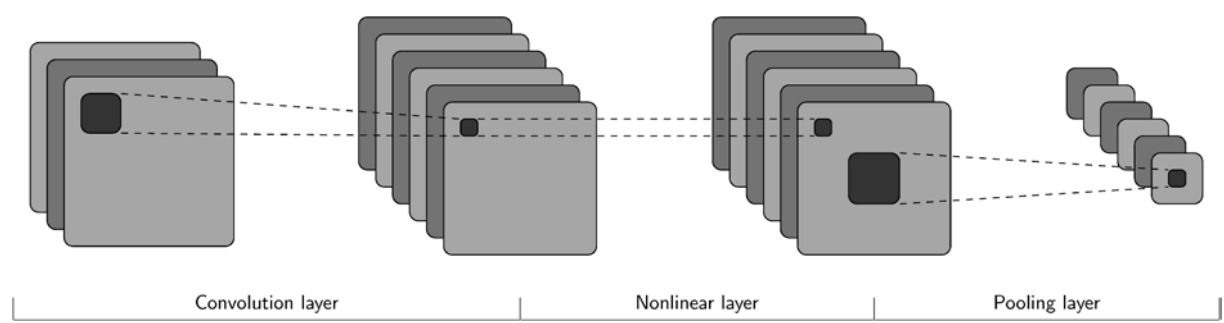

[FIGURE 6]

CNNs are designed to process multiple data types, especially two-dimensional images, and are directly inspired by the visual cortex of the brain. In the visual cortex, there is a hierarchy of two basic cell types: simple cells and complex cells [69]. Simple cells react to primitive patterns in sub-regions of visual stimuli, and complex cells synthesize the information from simple cells to identify more intricate forms. Since the visual cortex is such a powerful and natural visual processing system, CNNs are applied to imitate three key ideas: local connectivity, invariance to location, and invariance to local transition [8].

The basic structure of CNNs consists of convolution layers, nonlinear layers, and pooling layers (Figure 6). To use highly correlated sub-regions of data, groups of local weighted sums, called feature maps, are obtained at each convolution layer by computing convolutions between local patches and weight vectors called filters. Furthermore, since identical patterns can appear regardless of the location in the data, filters are applied repeatedly across the entire dataset, which also improves training efficiency by reducing the number of parameters to learn. Then nonlinear layers increase the nonlinear properties of feature maps. At each pooling layer, maximum or average subsampling of non-overlapping regions in feature maps is performed. This non-overlapping subsampling enables CNNs to handle somewhat different but semantically similar features and thus aggregate local features to identify more complex features.

Currently, CNNs are one of the most successful deep learning architectures owing to their outstanding capacity to analyze spatial information. Thanks to their developments in the field of object recognition, we believe the primary research achievements in bioinformatics will come from the biomedical imaging domain. Despite the different data characteristics between normal and biomedical imaging, CNN will nonetheless offer straightforward applications compared to other domains. Indeed, CNNs also have great potential in omics and biomedical signal processing. The three keys ideas of CNNs can be applied not only in a one-dimensional grid to discover meaningful recurring patterns with small variance, such as genomic sequence motifs, but also in two-dimensional grids, such as interactions within omics data and in time-frequency matrices of biomedical signals. Thus, we believe the popularity and promise of CNNs in bioinformatics applications will continue in the years ahead.

# Recurrent neural networks

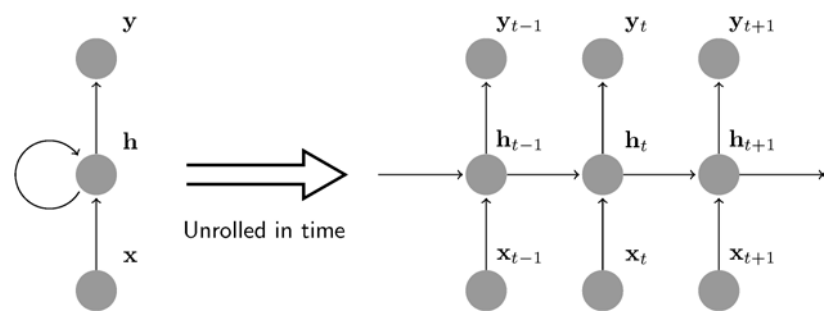

[FIGURE 7]

RNNs, which are designed to utilize sequential information, have a basic structure with a cyclic connection (Figure 7). Since input data are processed sequentially, recurrent computation is performed in the hidden units where cyclic connection exists. Therefore, past information is implicitly stored in the hidden units called state vectors, and output for the current input is computed considering all previous inputs using these state vectors [8]. Since there are many cases where both past and future inputs affect output for the current input (*e.g.,* in speech recognition), bidirectional recurrent neural networks (BRNNs) [70] have also been designed and used widely (Figure 8).

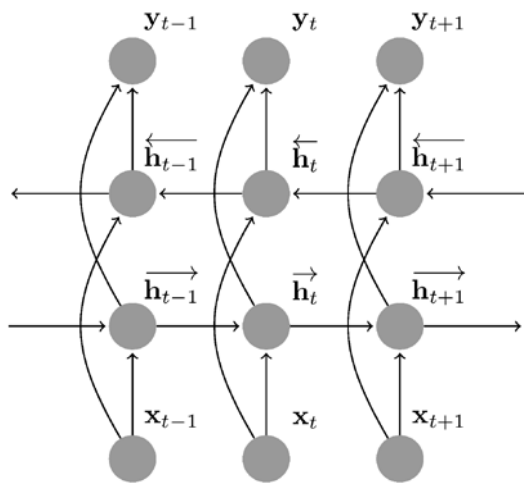

[FIGURE 8]

Although RNNs do not seem to be deep as DNNs or CNNs in terms of the number of layers, they can be regarded as an even deeper structure if unrolled in time (Figure 7). Therefore, for a long time, researchers struggled against vanishing gradient problems while training RNNs, and learning long-term dependency among data was difficult [35]. Fortunately, substituting the simple perceptron hidden units with more complex units such as LSTM [36, 37] or GRU [19], which function as memory cells, significantly helps to prevent the problem. More recently, RNNs have been used successfully in many areas including natural language processing [16, 17] and language translation [18, 19].

Even though RNNs have been explored less than DNNs and CNNs, they still provide very powerful analysis methods for sequential information. Since omics data and biomedical signals are typically sequential and often considered languages of nature, the capabilities of RNNs for mapping a variable-length input sequence to another sequence or fixed-size prediction are promising for bioinformatics research. With regard to biomedical imaging, RNNs are currently not the first choice of many researchers. Nevertheless, we believe that dissemination of dynamic CT and MRI [71, 72] would lead to the incorporation of RNNs and CNNs and elevate their importance in the long term. Furthermore, we expect that their successes in natural language processing will lead RNNs to be applied in biomedical text analysis [73] and that employing an attention mechanism [74-77] will improve performance and extract more relevant information from bioinformatics data.

# Emergent architectures

Emergent architectures refer to deep learning architectures besides DNNs, CNNs, and RNNs. In this review, we introduce three emergent architectures (*i.e.,* DST-NNs, MD-RNNs, and CAEs) and their applications in bioinformatics.

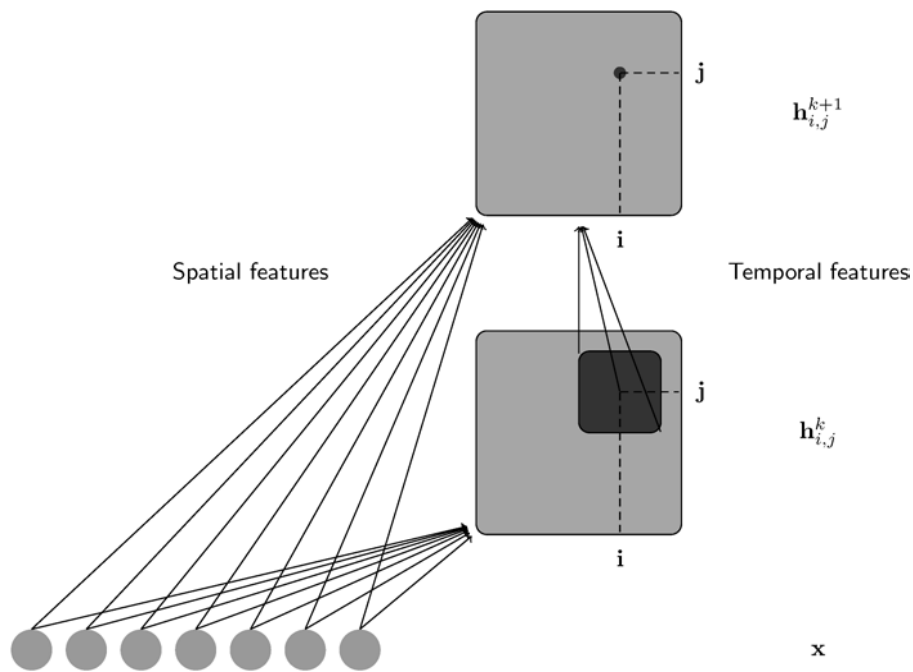

[FIGURE 9]

DST-NNs [38] are designed to learn multi-dimensional output targets through progressive refinement. The basic structure of DST-NNs consists of multi-dimensional hidden layers (Figure 9). The key aspect of the structure, progressive refinement, considers local correlations and is performed via input feature compositions in each layer: spatial features and temporal features. Spatial features refer to the original inputs for the whole DST-NN and are used identically in every layer. However, temporal features are gradually altered so as to progress to the upper layers. Except for the first layer, to compute each hidden unit in the current layer, only the adjacent hidden units of the same coordinate in the layer below are used so that local correlations are reflected progressively.

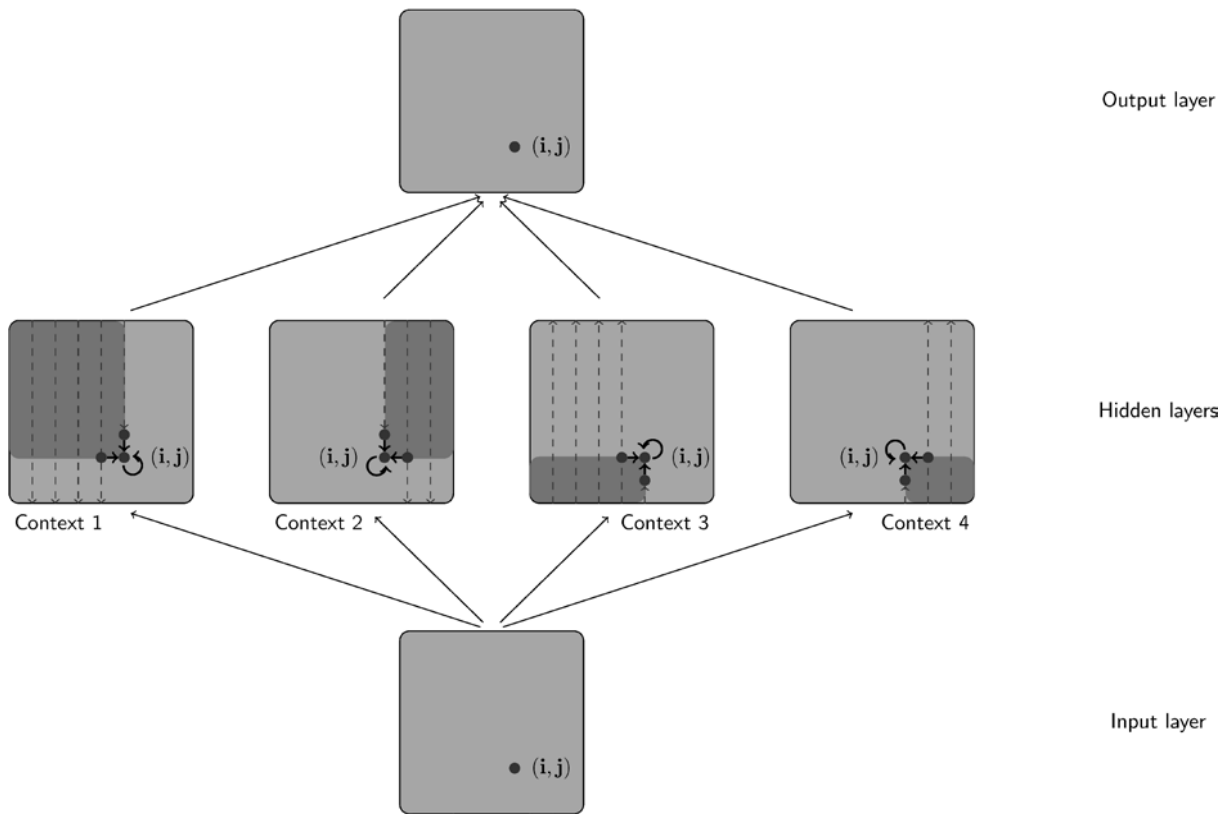

[FIGURE 10]

MD-RNNs [39] are designed to apply the capabilities of RNNs to non-sequential multi-dimensional data by treating them as groups of sequential data. For instance, two-dimensional data are treated as groups of horizontal and vertical sequence data. Similar to BRNNs which use contexts in both directions in one-dimensional data, MD-RNNs use contexts in all possible directions in the multi-dimensional data (Figure 10). In the example of a two-dimensional dataset, four contexts that vary with the order of data processing are reflected in the computation of four hidden units for each position in the hidden layer. The hidden units are connected to a single output layer, and the final results are computed with consideration of all possible contexts.

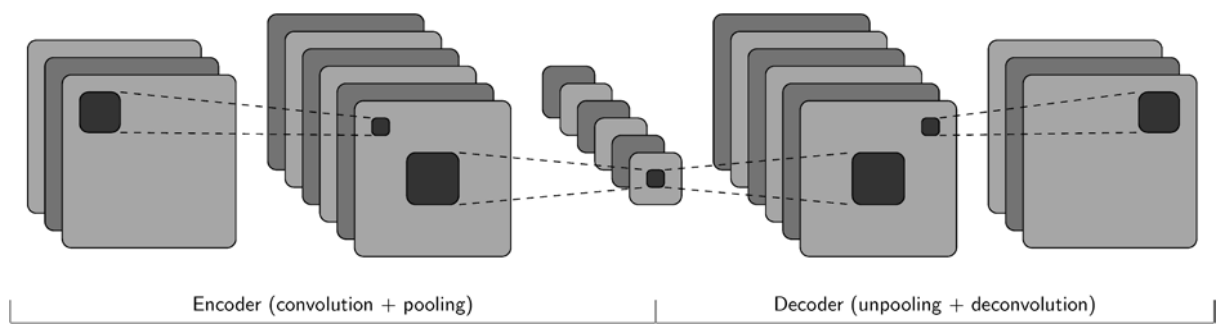

[FIGURE 11]

CAEs [40, 41] are designed to utilize the advantages of both AE and CNNs so that it can learn good hierarchical representations of data reflecting spatial information and be well regularized by unsupervised training (Figure 11). In training of AEs, reconstruction error is minimized using an encoder and decoder, which extract feature vectors from input data and recreate the data from the feature vectors, respectively. In CNNs, convolution and pooling layers can be regarded as a type of encoder. Therefore, the CNN encoder and decoder consisting of deconvolution and unpooling layers are integrated to form a CAE and are trained in the same manner as in AE.

Deep learning is a rapidly growing research area, and a plethora of new deep learning architecture is being proposed but awaits wide applications in bioinformatics. Newly proposed architectures have different advantages from existing architectures, so we expect them to produce promising results in various research areas. For example, the progressive refinement of DST-NNs fits the dynamic folding process of proteins and can be effectively utilized in protein structure prediction [38]; the capabilities of MD-RNNs are suitable for segmentation of biomedical images since segmentation requires interpretation of local and global contexts; the unsupervised representation learning with consideration of spatial information in CAEs can provide great advantages in discovering recurring patterns in limited and imbalanced bioinformatics data.

**TABLE 4**: Deep learning applied bioinformatics research avenues and input data

| | Input data | Research avenues |
|---|---|---|
| **Omics** | sequencing data (DNA-seq, RNA-seq, ChIP-seq, DNase-seq)<br>features from genomic sequence<br>    position specific scoring matrix (PSSM)<br>    physicochemical properties (steric parameter, volume)<br>    Atchley factors (FAC)<br>    1-dimensional structural properties<br>contact map (distance of amino acid pairs in 3D structure)<br>microarray gene expression | **Protein structure prediction [84-92]**<br>    1-dimensional structural properties<br>    contact map<br>    structure model quality assessment<br>**Gene expression regulation [93-107]**<br>    splice junction<br>    genetic variants affecting splicing<br>    sequence specificity<br>**Protein classification [108-110]**<br>    super family<br>    subcellular localization<br>**Anomaly classification [111]**<br>    Cancer |
| **Biomedical imaging** | magnetic resonance image (MRI)<br>radiographic image<br>positron emission tomography (PET)<br>histopathology image<br>volumetric electron microscopy image<br>retinal image<br>in situ hybridization (ISH) image | **Anomaly classification [122-132]**<br>    gene expression pattern<br>    cancer<br>    Alzheimer's disease<br>    schizophrenia<br>**Segmentation [133-141]**<br>    cell structure<br>    neuronal structure<br>    vessel map<br>    brain tumor<br>**Recognition [142-147]**<br>    cell nuclei<br>    finger joint<br>    anatomical structure<br>**Brain decoding [149-150]**<br>    behavior |
| **Biomedical signal processing** | ECoG, ECG, EMG, EOG<br>EEG (raw, wavelet, frequency, differential entropy)<br>extracted features from EEG<br>    normalized decay<br>    peak variation | **Brain decoding [158-170]**<br>    behavior<br>    emotion<br>**Anomaly classification [171-178]**<br>    Alzheimer's disease<br>    seizure<br>    sleep stage |

# Omics

In omics research, genetic information such as genome, transcriptome, and proteome data is used to approach problems in bioinformatics. Some of the most common input data in omics are raw biological sequences (*i.e.,* DNA, RNA, amino acid sequences) which have become relatively affordable and easy to obtain with next-generation sequencing technology. In addition, extracted features from sequences such as a position specific scoring matrices (PSSM) [78], physicochemical properties [79, 80], Atchley factors [81], and one-dimensional structural properties [82, 83] are often used as inputs for deep learning algorithms to alleviate difficulties from complex biological data and improve results. In addition, protein contact maps, which present distances of amino acid pairs in their three-dimensional structure, and microarray gene expression data are also used according to the characteristics of interest. We categorized the topics of interest in omics into four groups (Table 4). One of the most researched problems is protein structure prediction, which aims to predict the secondary structure or contact map of a protein [84-92]. Gene expression regulation [93-107], including splice junctions or RNA binding proteins, and protein classification [108-110], including super family or subcellular localization, are also actively investigated. Furthermore, anomaly classification [111] approaches have been used with omics data to detect cancer.

*Deep neural networks*

DNNs have been widely applied in protein structure prediction [84-87] research. Since complete prediction in three-dimensional space is complex and challenging, several studies have used simpler approaches, such as predicting the secondary structure or torsion angles of protein. For instance, Heffernan et al. [85] applied SAE to protein amino acid sequences to solve prediction problems for secondary structure, torsion angle, and accessible surface area. In another study, Spencer et al. [86] applied DBN to amino acid sequences along with PSSM and Atchley factors to predict protein secondary structure. DNNs have also shown great capabilities in the area of gene expression regulation [93-98]. For example, Lee et al. [94] utilized DBN in splice junction prediction, a major research avenue in understanding gene expression [112], and proposed a new DBN training method called boosted contrastive divergence for imbalanced data and a new regularization term for sparsity of DNA sequences; their work showed not only significantly improved performance but also the ability to detect subtle non-canonical splicing signals. Moreover, Chen et al. [96] applied MLP to both microarray and RNA-seq expression data to infer expression of up to 21000 target genes from only 1000 landmark genes. In terms of protein classification, Asgari et al. [108] adopted the skip-gram model, a widely known method in natural language processing, that can be considered a variant of MLP, and showed that it could effectively learn a distributed representation of biological sequences with general use for many omics applications, including protein family classification. For anomaly classification, Fakoor et al. [111] used principal component analysis (PCA) [113] to reduce the dimensionality of microarray gene expression data and applied SAE to classify various cancers, including acute myeloid leukemia, breast cancer, and ovarian cancer.

*Convolutional neural networks*

Relatively few studies have used CNNs to solve problems involving biological sequences, specifically gene expression regulation problems [99-104]; nevertheless, those have introduced the strong advantages of CNNs, showing their great promise for future research. First, an initial convolution layer can powerfully capture local sequence patterns and can be considered a motif detector for which PSSMs are solely learned from data instead of hard-coded. The depth of CNNs enables learning more complex patterns and can capture longer motifs, integrate cumulative effects of observed motifs, and eventually learn sophisticated regulatory codes [114]. Moreover, CNNs are suited to exploit the benefits of multitask joint learning. By training CNNs to simultaneously predict closely related factors, features with predictive strengths are more efficiently learned and shared across different tasks.

For example, as an early approach, Denas et al. [99] preprocessed ChIP-seq data into a two-dimensional matrix with the rows as transcription factor activity profiles for each gene and exploited a two-dimensional CNN similar to its use in image processing. Recently, more studies focused on directly using one-dimensional CNNs with biological sequence data. Alipanahi et al. [100] and Kelley et al. [103] proposed CNN-based approaches for transcription factor binding site prediction and 164 cell-specific DNA accessibility multitask prediction, respectively; both groups presented downstream applications for disease-associated genetic variant identification. Furthermore, Zeng et al. [102] performed a systematic exploration of CNN architectures for transcription factor binding site prediction and showed that the number of convolutional filters is more important than the number of layers for motif-based tasks. Zhou et al. [104] developed a CNN-based algorithmic framework, DeepSEA, that performs multitask joint learning of chromatin factors (i.e., transcription factor binding, DNase I sensitivity, histone-mark profile) and prioritizes expression quantitative trait loci and disease-associated genetic variants based on the predictions.

*Recurrent neural networks*

RNNs are expected to be an appropriate deep learning architecture because biological sequences have variable lengths, and their sequential information has great importance. Several studies have applied RNNs to protein structure prediction [88-90], gene expression regulation [105-107], and protein classification [109, 110]. In early studies, Baldi et al. [88] used BRNNs with perceptron hidden units in protein secondary structure prediction. Thereafter, the improved performance of LSTM hidden units became widely recognized, so Sønderby et al. [110] applied BRNNs with LSTM hidden units and a one-dimensional convolution layer to learn representations from amino acid sequences and classify the subcellular locations of proteins. Furthermore, Park et al. [105] and Lee et al. [107] exploited RNNs with LSTM hidden units in microRNA identification and target prediction and obtained significantly improved accuracy relative to state-of-the-art approaches demonstrating the high capacity of RNNs to analyze biological sequences.

*Emergent architectures*

Emergent architectures have been used in protein structure prediction research [91, 92], specifically in contact map prediction. Di Lena et al. [91] applied DST-NNs using spatial features including protein secondary structure, orientation probability, and alignment probability. Additionally, Baldi et al. [92] applied MD-RNNs to amino acid sequences, correlated profiles, and protein secondary structures.

# Biomedical imaging

Biomedical imaging [115] is another an actively researched domain with wide application of deep learning in general image-related tasks. Most biomedical images used for clinical treatment of patients—magnetic resonance imaging (MRI) [116, 117], radiographic imaging [118, 119], positron emission tomography (PET) [120], and histopathology imaging [121]—have been used as input data for deep learning algorithms. We categorized the research avenues in biomedical imaging into four groups (Table 4). One of the most researched problems is anomaly classification [122-132] to diagnose diseases such as cancer or schizophrenia. As in general image-related tasks, segmentation [133-141] (*i.e.,* partitioning specific structures such as cellular structures or a brain tumor) and recognition [142-147] (*i.e.,* detection of cell nuclei or a finger joint) are studied frequently in biomedical imaging. Studies of popular high content screening [148], which involves quantifying microscopic images for cell biology, are covered in the former groups [128, 134, 137]. Additionally, cranial MRIs have been used in brain decoding [149, 150] to interpret human behavior or emotion.

*Deep neural networks*

In terms of biomedical imaging, DNNs have been applied in several research areas, including anomaly classification [122-124], segmentation [133], recognition [142, 143], and brain decoding [149, 150]. Plis et al. [122] classified schizophrenia patients from brain MRIs using

DBN, and Xu et al. [142] used SAE to detect cell nuclei from histopathology images. Interestingly, similar to handwritten digit image recognition, Van Gerven et al. [149] classified handwritten digit images with DBN not by analyzing the images themselves but by indirectly analyzing indirectly functional MRIs of participants who are looking at the digit images.

*Convolutional neural networks*

The largest number of studies have been conducted in biomedical imaging, since these avenues are similar to general image-related tasks. In anomaly classification [125-132], Roth et al. [125] applied CNNs to three different CT image datasets to classify sclerotic metastases, lymph nodes, and colonic polyps. Additionally, Ciresan et al. [128] used CNNs to detect mitosis in breast cancer histopathology images, a crucial approach for cancer diagnosis and assessment. PET images of esophageal cancer were used by Ypsilantis et al. [129] to predict responses to neoadjuvant chemotherapy. Other applications of CNNs can be found in segmentation [134-140] and recognition [144-147]. For example, Ning et al. [134] studied pixel-wise segmentation patterns of the cell wall, cytoplasm, nuclear membrane, nucleus, and outside media using microscopic image, and Havaei et al. [139] proposed a cascaded CNN architecture exploiting both local and global contextual features and performed brain tumor segmentation from MRIs. For recognition, Cho et al. [144] researched anatomical structure recognition among CT images, and Lee et al. [145] proposed a CNN-based finger joint detection system, FingerNet, which is a crucial step for medical examinations of bone age, growth disorders, and rheumatoid arthritis [151].

*Recurrent neural networks*

Traditionally, images are considered data that involve internal correlations or spatial information rather than sequential information. Treating biomedical images as non-sequential data, most studies in biomedical imaging have chosen approaches involving DNNs or CNNs instead of RNNs.

*Emergent architectures*

Attempts to apply the unique capabilities of RNNs to image data using augmented RNN structures have continued. MD-RNNs [39] have been applied beyond two-dimensional images to three-dimensional images. For example, Stollenga et al. [141] applied MD-RNNs to three-dimensional electron microscopy images and MRIs to segment neuronal structures.

# Biomedical signal processing

Biomedical signal processing [115] is a domain where researchers use recorded electrical activity from the human body to solve problems in bioinformatics. Various data from EEG [152], electrocorticography (ECoG) [153], electrocardiography (ECG) [154], electromyography (EMG) [155], and electrooculography (EOG) [156, 157] have been used, with most studies focusing on EEG activity so far. Because recorded signals are usually noisy and include many artifacts, raw signals are often decomposed into wavelet or frequency components before they are used as input in deep learning algorithms. In addition, human-designed features like normalized decay and peak variation are used in some studies to improve the results. We categorized the research avenues in biomedical signal processing into two groups (Table 4): brain decoding [158-170] using EEG signals and anomaly classification [171-178] to diagnose diseases.

*Deep neural networks*

Since biomedical signals usually contain noise and artifacts, decomposed features are more frequently used than raw signals. In brain decoding [158-163], An et al. [159] applied DBN to the frequency components of EEG signals to classify left- and right-hand motor imagery skills. Moreover, Jia et al. [161] and Jirayucharoensak et al. [163] used DBN and SAE, respectively, for emotion classification. In anomaly classification [171-175], Huanhuan et al. [171] published one of the few studies applying DBN to ECG signals and classified each beat into either a normal or abnormal beat. A few studies have used raw EEG signals. Wulsin et al. [172] analyzed individual second-long waveform abnormalities using DBN with both raw EEG signals and extracted features as inputs, whereas Zhao et al. [174] used only raw EEG signals as inputs for DBN to diagnose Alzheimer's disease.

*Convolutional neural networks*

Raw EEG signals have been analyzed in brain decoding [164-167] and anomaly classification [176] via CNNs, which perform one-dimensional convolutions. For instance, Stober et al. [165] classified the rhythm type and genre of music that participants listened to, and Cecotti et al. [167] classified characters that the participants viewed. Another approach to apply CNNs to biomedical signal processing was reported by Mirowski et al. [176], who extracted features such as phase-locking synchrony and wavelet coherence and coded them as pixel colors to formulate two-dimensional patterns. Then, ordinary two-dimensional CNNs, like the one used in biomedical imaging, were used to predict seizures.

*Recurrent neural networks*

Since biomedical signals represent naturally sequential data, RNNs are an appropriate deep learning architecture to analyze data and are expected to produce promising results. To present some of the studies in brain decoding [168] and anomaly classification [177, 178], Petrosian et al. [177] applied perceptron RNNs to raw EEG signals and corresponding wavelet decomposed features to predict seizures. In addition, Davidson et al. [178] used LSTM RNNs on EEG log-power spectra features to detect lapses.

*Emergent architectures*

CAE has been applied in a few brain decoding studies [169, 170]. Wang et al. [169] performed finger flex and extend classifications using raw ECoG signals. In addition, Stober et al. [170] classified musical rhythms that participants listened to with raw EEG signals.

# Discussion

## Limited and imbalanced data

Considering the necessity of optimizing a tremendous number of weight parameters in neural networks, most deep learning algorithms have assumed sufficient and balanced data. Unfortunately, however, this is usually not true for problems in bioinformatics. Complex and expensive data acquisition processes limit the size of bioinformatics datasets. In addition, such processes often show significantly unequal class distributions, where an instance from one class is significantly higher than instances from other classes [179]. For example in clinical or disease-related cases, there is inevitably less data from treatment groups than from the normal (control) group. The former are also rarely disclosed to the public due to privacy restrictions and ethical requirements creating a further imbalance in available data [180].

A few assessment metrics have been used to clearly observe how limited and imbalanced data might compromise the performance of deep learning [181]. While accuracy often gives misleading results, the F-measure, the harmonic mean of precision and recall, provides more insightful performance scores. To measure performance over different class distributions, the area under the receiver operating characteristic curve (AUC) and the area under the precision-recall curve (AUC-PR) are commonly used. These two measures are strongly correlated such that a curve dominates in one measure if and only if it dominates in the other. Nevertheless, in contrast with AUC-PR, AUC might present a more optimistic view of performance, since false positive rates in the receiver operating characteristic curve fail to capture large changes of false positives if classes are negatively skewed [182].

Solutions to limited and imbalanced data can be divided into three major groups [181, 183] : data preprocessing, cost-sensitive learning and algorithmic modification. Data preprocessing typically provides a better dataset through sampling or basic feature extraction. Sampling methods balance the distribution of imbalanced data, and several approaches have been proposed, including informed undersampling [184], the synthetic minority oversampling technique [185], and cluster-based sampling [186]. For example, Li et al. [127] and Roth et al. [146] performed enrichment analyses of CT images through spatial deformations such as random shifting and rotation. Although basic feature extraction methods deviate from the concept of deep learning, they are occasionally used to lessen the difficulties of learning from limited and imbalanced data. Research in bioinformatics using human designed features as input data such as PSSM from genomics sequences or wavelet energy from EEG signals can be understood in the same context [86, 92, 172, 176].

Cost-sensitive learning methods define different costs for misclassifying data examples from individual classes to solve the limited and imbalanced data problems. Cost sensitivity can be applied in an objective loss function of neural networks either explicitly or implicitly [187]. For example, we can explicitly replace the objective loss function to reflect class imbalance or implicitly modify the learning rates according to data instance classes during training.

Algorithmic modification methods accommodate learning algorithms to increase their suitability for limited and imbalanced data. A simple and effective approach is adoption of pre-training. Unsupervised pre-training can be a great help to learn representation for each class and to produce more regularized results [68]. In addition, transfer learning, which consists of pre-training with sufficient data from similar but different domains and fine-tuning with real data, has great advantages [24, 188]. For instance, Lee et al. [107] proposed a microRNA target prediction method, which exploits unsupervised pre-training with RNN based AE, and achieved a >25% increase in F-measure compared to the existing alternatives. Bar et al. [132]

performed transfer learning using natural images from the ImageNet database [189] as pre-training data and fine-tuned with chest X-ray images to identify chest pathologies and to classify healthy and abnormal images. In addition to pre-training, sophisticated training methods have also been executed. Lee et al. [94] suggested DBN with boosted categorical RBM, and Havaei et al. [139] suggested CNNs with two-phase training, combining ideas of undersampling and pre-training.

**Changing the black-box into the white-box**

A main criticism against deep learning is that it is used as a black-box: even though it produces outstanding results, we know very little about how such results are derived internally. In bioinformatics, particularly in biomedical domains, it is not enough to simply produce good outcomes. Since many studies are connected to patients' health, it is crucial to change the black-box into the white-box providing logical reasoning just as clinicians do for medical treatments.

Transformation of deep learning from the black-box into the white-box is still in the early stages. One of the most widely used approaches is interpretation through visualizing a trained deep learning model. In terms of image input, a deconvolutional network has been proposed to reconstruct and visualize hierarchical representations for a specific input of CNNs [190]. In addition, to visualize a generalized class representative image rather than being dependent on a particular input, gradient ascent optimization in input space through backpropagation-to-input (*cf.* backpropagation-to-weights) has provided another effective methodology [191, 192]. Regarding genomic sequence input, several approaches have been proposed to infer PSSMs from a trained model and to visualize the corresponding motifs with heat maps or sequence logos. For example, Lee et al. [94] extracted motifs by choosing the most class discriminative weight vector among those in the first layer of DBN; DeepBind [100] and DeMo [101] extracted motifs from trained CNNs by counting nucleotide frequencies of positive input subsequences with high activation values and backpropagation-to-input for each feature map, respectively.

Specifically for transcription factor binding site prediction, Alipanahi et al. [100] developed a visualization method, a mutation map, for illustrating the effects of genetic variants on binding scores predicted by CNNs. A mutation map consists of a heat map, which shows how much each mutation alters the binding score, and the input sequence logo, where the height of each base is scaled as the maximum decrease of binding score among all possible mutations.

Moreover, Kelley et al. [103] further complemented the mutation map with a line plot to show the maximum increases as well as the maximum decreases of prediction scores. In addition to interpretation through visualization, attention mechanisms [74-77] designed to focus explicitly on salient points and the mathematical rationale behind deep learning [193, 194] are being studied.

**Selection of an appropriate deep learning architecture and hyperparameters**

Choosing the appropriate deep learning architecture is crucial to proper applications of deep learning. To obtain robust and reliable results, awareness of the capabilities of each deep learning architecture and selection according to capabilities in addition to input data characteristics and research objectives are essential. However, to date, the advantages of each architecture are only roughly understood; for example, DNNs are suitable for analysis of internal correlations in high-dimensional data, CNNs are suitable for analysis of spatial information, and RNNs are suitable for analysis of sequential information [7]. Indeed, a detailed methodology for selecting the most appropriate or "best fit" deep learning architecture remains a challenge to be studied in the future.

Even once a deep learning architecture is selected, there are many hyperparameters—the number of layers, the number of hidden units, weight initialization values, learning iterations, and even the learning rate—for researchers to set, all of which can influence the results remarkably [195]. For many years, hyperparameter tuning was rarely systematic and left up to human machine learning experts. Nevertheless, automation of machine learning research, which aims to automatically optimize hyperparameters is growing constantly [196]. A few algorithms have been proposed including sequential model based global optimization [197], Bayesian optimization with Gaussian process priors [198], and random search approaches [199].

**Multimodal deep learning**

Multimodal deep learning [200], which exploits information from multiple input sources, is a promising avenue for the future of deep learning research. In particular, bioinformatics is expected to benefit greatly, as it is a field where various types of data can be assimilated naturally [201]. For example, not only are omics data, images, signals, drug responses, and electronic medical records available as input data, but X-ray, CT, MRI, and PET forms are also available from a single image.

A few bioinformatics studies have already begun to use multimodal deep learning. For example, Suk et al. [124] studied Alzheimer's disease classification using cerebrospinal fluid and brain images in the forms of MRI and PET scan and Soleymani et al. [168] conducted an emotion detection study with both EEG signal and face image data.

**Accelerating deep learning**

As more deep learning model parameters and training data become available, better learning performances can be achieved. However, at the same time, this inevitably leads to a drastic increase in training time, emphasizing the necessity for accelerated deep learning [7, 25].

Approaches to accelerating deep learning can be divided into three groups: advanced optimization algorithms, parallel and distributed computing, and specialized hardware. Since the main reason for long training times is that parameter optimization through plain SGD takes too long, several studies have focused on advanced optimization algorithms [202]. To this end, some widely employed algorithms include Adagrad [48], Adam [49], batch normalization [55], and Hessian-free optimization [203]. Parallel and distributed computing can significantly accelerate the time to completion and have enabled many deep learning studies [204-208]. These approaches exploit both scale-up methods, which use a graphic processing unit, and scale-out methods, which use large-scale clusters of machines in a distributed environment. A few deep learning frameworks, including the recently released DeepSpark [209] and TensorFlow [210] provide parallel and distributed computing abilities. Although development of specialized hardware for deep learning is still in its infancy, it will provide major accelerations and become far more important in the long term [211]. Currently, field programmable gate array based processors are under development, and neuromorphic chips modeled from the brain are greatly anticipated as promising technologies [212-214].

**Future trends of deep learning**

Incorporation of traditional deep learning architectures is a promising future trend. For instance, joint networks of CNNs and RNNs integrated with attention models have been applied in image captioning [75], video summarization [215], and image question answering [216]. A few studies toward augmenting the structures of RNNs have been conducted as well. Neural Turing machines [217] and memory networks [218] have adopted addressable external memory in RNNs and shown great results for tasks requiring intricate inferences, such as algorithm learning and complex question answering. Recently, adversarial examples, which degrade performance with small human-imperceptible perturbations, have received increased attention from the machine learning community [219, 220]. Since adversarial training of neural networks can result in regularization to provide higher performance, we expect additional studies in this area, including those involving adversarial generative networks [221] and manifold regularized networks [222].

In terms of learning methodology, semi-supervised learning and reinforcement learning are also receiving attention. Semi-supervised learning exploits both unlabeled and labeled data, and a few algorithms have been proposed. For example, ladder networks [223] add skip connections to MLP or CNNs, and simultaneously minimize the sum of supervised and unsupervised cost functions to denoise representations at every level of the model. Reinforcement learning leverages reward outcome signals resulting from actions rather than correctly labeled data. Since reinforcement learning most closely resembles how humans actually learn, this approach has great promise for artificial general intelligence [224]. Currently, its applications are mainly focused on game playing [4] and robotics [225].

# Conclusion

As we enter the major era of big data, deep learning is taking center stage for international academic and business interests. In bioinformatics, where great advances have been made with conventional machine learning, deep learning is anticipated to produce promising results. In this review, we provided an extensive review of bioinformatics research applying deep learning in terms of input data, research objectives, and the characteristics of established deep learning architectures. We further discussed limitations of the approach and promising directions of future research.

Although deep learning holds promise, it is not a silver bullet and cannot provide great results in ad hoc bioinformatics applications. There remain many potential challenges, including limited or imbalanced data, interpretation of deep learning results, and selection of an appropriate architecture and hyperparameters. Furthermore, to fully exploit the capabilities of deep learning, multimodality and acceleration of deep learning require further study. Thus, we are confident that prudent preparations regarding the issues discussed herein are key to the success of future deep learning approaches in bioinformatics. We believe that this review will provide valuable insight and serve as a starting point for application of deep learning to advance bioinformatics in future research.

## Funding

This work was supported by the National Research Foundation (NRF) of Korea grants funded by the Korean Government (Ministry of Science, ICT and Future Planning) [No. 2011-0009963, No. 2014M3C9A3063541]; the ICT R&D program of MSIP/ITP [14-824-09-014, Basic Software Research in Human-level Lifelong Machine Learning (Machine Learning Center)]; and SNU ECE Brain Korea 21+ project in 2016.

## Acknowledgements

The authors would like to thank Prof. Russ Altman and Prof. Tsachy Weissman at Stanford University, Prof. Honglak Lee at University of Michigan, Prof. V. Narry Kim and Prof. Daehyun Baek at Seoul National University, and Prof. Young-Han Kim at University of California, San Diego for helpful discussions on applying artificial intelligence and machine learning to bioinformatics.

## Figure captions

**Figure 1:** Approximate number of published deep learning articles by year. The number of articles is based on the search results on http://www.scopus.com with the two queries: "Deep learning," "Deep learning" AND "bio*".

**Figure 2:** Application of deep learning in bioinformatics research. (A) Overview diagram with input data and research objectives. (B) A research example in the omics domain. Prediction of splice junctions in DNA sequence data with a deep neural network [94]. (C) A research example in biomedical imaging. Finger joint detection from X-ray images with a convolutional neural network [145]. (D) A research example in biomedical signal processing. Lapse detection from EEG signal with a recurrent neural network [178].

**Figure 3:** Relationships and high-level schematics of artificial intelligence, machine learning, representation learning, and deep learning [7].

**Figure 4:** Basic structure of DNNs with input units x, three hidden units $h_1$, $h_2$, and $h_3$, in each layer and output units y [26]. At each layer, the weighted sum and nonlinear function of its inputs are computed so that the hierarchical representations can be obtained.

**Figure 5:** Unsupervised layer-wise pre-training process in SAE and DBN [29]. First, weight vector W1 is trained between input units x and hidden units h$_1$ in the first hidden layer as an RBM or AE. After the W1 is trained, another hidden layer is stacked, and the obtained representations in h$_1$ are used to train W2 between hidden units h$_1$ and h$_2$ as another RBM or AE. The process is repeated for the desired number of layers.

**Figure 6:** Basic structure of CNNs consisting of a convolution layer, a nonlinear layer, and a pooling layer [32]. The convolution layer of CNNs uses multiple learned filters to obtain multiple filter maps detecting low-level filters, and then the pooling layer combines them into higher-level features.

**Figure 7:** Basic structure of RNNs with an input unit x, a hidden unit h, and an output unit y [8]. A cyclic connection exists so that the computation in the hidden unit receives inputs from the hidden unit at the previous time step and from the input unit at the current time step. The recurrent computation can be expressed more explicitly if the RNNs are unrolled in time. The index of each symbol represents the time step. In this way, h$_t$ receives input from x$_t$ and h$_{t-1}$ and then propagates the computed results to y$_t$ and h$_{t+1}$.

**Figure 8:** Basic structure of BRNNs unrolled in time [70]. There are two hidden units $\vec{h}_t$ and $\overleftarrow{h}_t$ for each time step. $\vec{h}_t$ receives input from x$_t$ and $\vec{h}_{t-1}$ to reflect past information; $\overleftarrow{h}_t$ receives input from x$_t$ and $\overleftarrow{h}_{t+1}$ to reflect future information. The information from both hidden units is propagated to y$_t$.

**Figure 9:** Basic structure of DST-NNs [38]. The notation $h_{i,j}^k$ represents the hidden unit at (i, j) coordinate of the k-th hidden layer. To conduct the progressive refinement, the neighborhood units of $h_{i,j}^k$ and input units x are used in the computation of $h_{i,j}^{k+1}$.

**Figure 10:** Basic structure of MD-RNNs for two-dimensional data [39]. There are four groups of two-dimensional hidden units, each reflecting different contexts. For example, the (i, j) hidden unit in context 1 receives input from the (i–1, j) and (i, j–1) hidden units in context 1 and the (i, j) unit from the input layer so that the upper-left information is reflected. The hidden units from all four contexts are propagated to compute the (i, j) unit in the output layer.

**Figure 11:** Basic structure of CAEs consisting of a convolution layer and a pooling layer working as an encoder and a deconvolution layer and an unpooling layer working as a decoder [41]. The basic idea is similar to the AE, which learns hierarchical representations through reconstructing its input data, but CAE additionally utilizes spatial information by integrating convolutions.